\documentclass[11pt,a4paper]{article}
\usepackage[hyperref]{emnlp2020}
\usepackage{times}
\usepackage{latexsym}

\usepackage{microtype}

\aclfinalcopy %

\title{Stay Hungry, Stay Focused: Generating Informative and Specific Questions in Information-Seeking Conversations}

\author{Peng Qi\qquad Yuhao Zhang \qquad Christopher D. Manning \\
Stanford University \\
Stanford, CA 94305 \\
  \texttt{\{pengqi, yuhaozhang, manning\}@stanford.edu} }

\date{}

\usepackage{xcolor}
\usepackage{mwe}
\usepackage{subfigure}

\usepackage{amsmath, amsfonts}
\usepackage{booktabs}

\usepackage{multirow}
\usepackage{array}

\usepackage[normalem]{ulem}

\usepackage{pgfplots}
\pgfplotsset{width=0.47\textwidth,compat=1.9, height=2in}

\definecolor{red}{HTML}{E31A1C}
\definecolor{blue}{HTML}{1F78B4}
\definecolor{green}{HTML}{33A02C}
\definecolor{orange}{HTML}{FF7F00}
\definecolor{purple}{HTML}{6A3D9A}

\newcommand{\ie}{\textit{i.e.}}
\newcommand{\eg}{\textit{e.g.}}

\newcommand{\fone}{F\textsubscript{1}}

\newcommand{\answer}[2]{{\color{blue}[\uwave{#2}]\textsubscript{Ans\textsubscript{#1}}}}

\newcommand{\squad}{SQuAD}
\newcommand{\quac}{QuAC}
\newcommand{\coqa}{CoQA}

\newcommand{\diff}[2]{{\color{red} #1}{\color{green} #2}}
\renewcommand{\diff}[2]{#2}

\begin{document}
\maketitle
\begin{abstract}

We investigate the problem of generating informative questions in information-asymmetric conversations.
Unlike previous work on question generation which largely assumes knowledge of what the answer might be, we are interested in the scenario where the questioner is not given the context from which answers are drawn, but must reason pragmatically about how to acquire new information, given the shared conversation history.
We identify two core challenges:  (1) formally defining the informativeness of potential questions, and (2) exploring the prohibitively large space of potential questions to find the good candidates.
To generate pragmatic questions, we use reinforcement learning to optimize an informativeness metric we propose, combined with a reward function designed to promote more specific questions.
We demonstrate that the resulting pragmatic questioner substantially improves the informativeness and specificity of questions generated over a baseline model, as evaluated by our metrics as well as humans.

\end{abstract}

\section{Introduction}

\begin{figure}[!ht]
    \centering
    \includegraphics[width=0.48\textwidth]{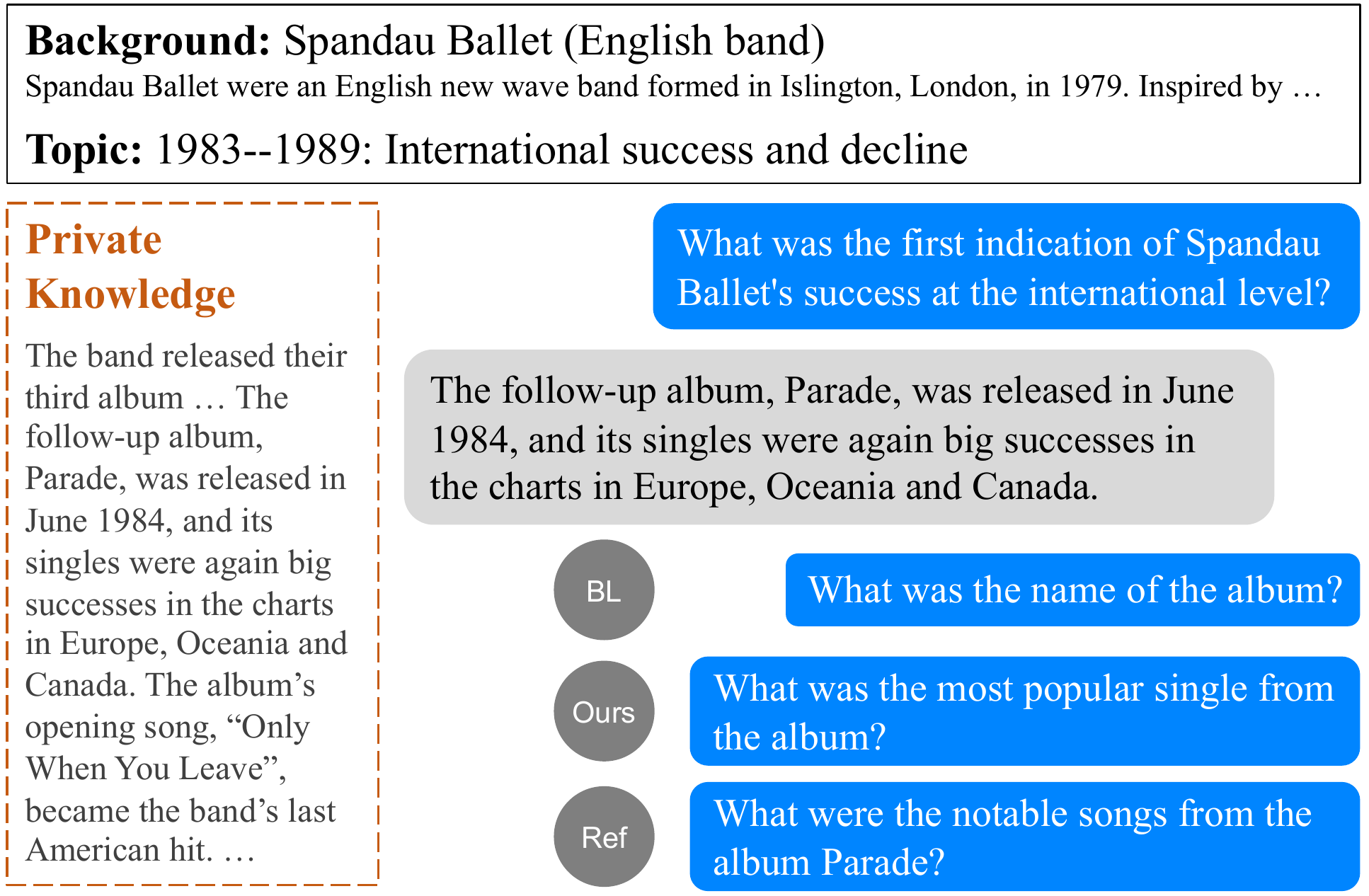}
    \caption{Asking questions in a conversation to acquire information.
    In this communication setting, the question asker has access to the background and topic, but no access to the private textual knowledge that contains the answer.
    In this example, the baseline non-pragmatic question generator (BL) generates an uninformative question (one that has already been answered), while our pragmatic system (Ours) and humans (Ref) actively seek new information.}
    \label{fig:example}
\end{figure}

Conversations are a primary means to seek and communicate information between humans, where asking the right question is an important prerequisite for effective exchange of knowledge.
Learning to ask questions in conversations can help computer systems not only acquire new knowledge, but also engage human interlocutors by making them feel heard \cite{huang2017doesnt}.

Previous work on question generation often falls into three classes:
generating questions according to a discrete schema or end goal \cite{bordes2017learning, zhang2018goal},
transforming the answer statement into a question \cite{mitkov2003computer,rus2010first,heilman2010good}, or generating questions with data-driven systems by conditioning on the context where the answer comes from \cite{du2017learning,zhou2017neural}.
Despite their successful adaptation to conversations to predict the question that elicits the observed answer \cite{gao2019interconnected,pan2019reinforced,nakanishi2019towards}, they are not suitable for modeling communication of knowledge in open-domain conversations, because the crucial problem of \emph{what to communicate} has already been assumed to be addressed by conditioning on the schema of information need or the context that contains the answer.

We instead study the problem of question generation in a more realistic setting, \ie, in \emph{open-domain information-seeking conversations} where the question asker cannot access the answering context.
This is an important step towards practical natural language processing (NLP) systems that can reason about the state of mind of agents they interact with purely through natural language interactions, so that they can generate more helpful responses.
In this paper, we build a question generator that reasons pragmatically about what information the answerer can provide, and generates questions to gather new information in a conversation (see Figure \ref{fig:example} for an example).

We identify several key challenges in this task: (1) generating informative questions without access to potential answers;
(2) evaluating generated questions beyond comparing them to the reference question, because multiple questions can reveal unseen information despite being very different to each other; (3) navigating a large search space of potential questions to improve informativeness by reasoning about the other agent's knowledge, which is more complex than limited reference games in previous work on computational pragmatics.

To address these issues, we first develop a baseline question generation model that generates questions in a conversation without conditioning on the unseen knowledge.
We then propose automatic metrics to quantify how much new information questions reveal, as well as how specific they are to the conversation.
Next, we use reinforcement learning to optimize our question generator on these metrics.
In our experiments on the \quac{} dataset, we show that the proposed method substantially improves the specificity and informativeness of the generated questions as evaluated by our automatic metrics.
These results are corroborated by blinded human evaluation, where  questions generated by our system are also of higher overall quality than those by the baseline system as judged by humans.
To recap, our main contributions are:\footnote{We release our code and models at \url{https://github.com/qipeng/stay-hungry-stay-focused}.}
\begin{itemize}
    \setlength\itemsep{0em}
    \item To the best of our knowledge, our work represents the first attempt at studying question generation to seek information in open-domain communication, which involves challenging NLP problems, \eg, evaluation of open-ended language generation and pragmatic reasoning;
    \item To address these problems, we propose automatic metrics to quantify the informativeness and specificity of questions, which are essential for efficient iterative system development;
    \item We show that optimizing the proposed metrics via reinforcement learning leads to a system that behaves pragmatically and has improved communication efficiency, as also verified by human evaluation.
    This represents a practical method for pragmatic reasoning in an open-domain communication setting.
\end{itemize}

\section{Related Work}

\paragraph{Question Generation.}
\diff{}{Question generation has long been studied in the education and psychology communities as a means to assess and promote reading comprehension in humans \cite{davey1986effects}.
In natural language processing,} question generation has been explored to improve the systems in various natural language processing tasks, \eg, the quality of question answering systems \cite{duan2017question} as well as information retrieval %
in an open-domain question answering system %
\cite{nogueira2019document}.

Some of the first question generation systems are rule-based \cite{mitkov2003computer, rus2010first,heilman2010good}, %
while large-scale question answering datasets, \eg, \squad\ \cite{rajpurkar2016squad,rajpurkar2018know}, have recently kindled research interest in data-driven approaches. %
\citet{du2017learning} and \citet{zhou2017neural} apply sequence-to-sequence (seq2seq) models to generate \squad{} questions from Wikipedia sentences containing the answers.

The release of large conversational question answering datasets such as \quac{} \cite{choi2018quac} and \coqa{} \cite{reddy2019coqa} enabled
\citet{gao2019interconnected}, \citet{pan2019reinforced}  and \citet{nakanishi2019towards} to extend previous neural seq2seq question generators by conditioning them on the conversation history and the context that contains the answer, while
\citet{scialom2019ask}    remove answers to the reference question to generate curiosity-driven questions from the rest of the context.

Despite their success, most existing approaches to question generation are limited to either \emph{reading comprehension} settings where potential answers are known \emph{a priori}, or \emph{goal-oriented settings} where the schema of knowledge is limited \cite{bordes2017learning,zhang2018goal}.
This %
prevents them from being applied to an \emph{open-domain communication} setting, where the purpose of questions is to acquire information that is unknown ahead of time.

\paragraph{Evaluating System-generated Questions.}
Automatic evaluation of system-generated text has long been an important topic in NLP.
Traditional $n$-gram overlap-based approaches \cite{papineni2002bleu, lin2004rouge} are computationally efficient, but have been shown to correlate poorly with human judgement of quality \cite{novikova2017need}.
More recently, \citet{zhang2020bertscore} leverage large pretrained language models \cite[BERT,][]{devlin2019bert} to relax the limitation of exact $n$-gram overlap.
\citet{hashimoto2019unifying} combine human judgement with system-reported likelihood of generated text to make population-level estimates of quality and diversity.
However, most existing metrics either evaluate generated text against very few references, or provide only relative ranking for multiple systems at a population level rather than reliable feedback for each example.
This renders them inapplicable to generating informative questions in a conversation, where multiple questions can be equally informative and relevant in a given scenario, and per-example feedback is necessary.

\paragraph{Pragmatic Reasoning for Informativeness.}
Pragmatic reasoning is tightly related to informativeness and efficiency in communication.
Starting from the cooperative maxims for conversational pragmatic reasoning \cite{grice1975logic}, \citet{frank2012predicting} developed a computational framework that has been applied to reference games with images \cite{andreas2016reasoning} and colors \cite{monroe2017colors}, as well as generating descriptions for images \cite{cohngordon2019incremental}. %
Decision-theoretic principles \cite{vanrooy2003questioning} have also been applied to quantify the informativeness of community questions %
\cite{rao2018learning}.
These approaches usually assume that either the list of \emph{referents} (images, colors, or answers) or the space of utterances (descriptions or questions) is enumerable or can be directly sampled from, or both.
More crucially, the speaker agent usually has complete access to this information to readily gauge the effect of different utterances.
We instead study a more realistic information-seeking setting, where the questioner cannot access the answers, let alone aggregate them for pragmatic reasoning, and where these simplifying assumptions will not hold.

\section{Method}

In this section, we outline the setup for the communication problem we set out to address, present a baseline system, and lay out our approach to extending it to reason pragmatically to acquire information more efficiently.

\subsection{Problem Setup}

We consider a communication game between two agents, a teacher and a student (see Figure \ref{fig:example} for an example).
The two agents share a common topic of discussion $\mathcal T$ (\emph{Background} and \emph{Topic} in the figure), as well as a common goal for the student to acquire some knowledge $\mathcal K$ on this topic that only the teacher has direct access to (\emph{Private Knowledge} in the figure).
We consider the scenario where the agents can only communicate to each other by engaging in a conversation, where the conversation history $\mathcal H$ is shared between the agents.
We further constrain the conversation to one where the student asks questions about the shared topic, and the teacher provides answers based on $\mathcal K$.
Note that this setup is very similar to that of the ``Game of Interrogation'' by \cite{groenendijk1999logic}, except we relax the definition, using natural language instead of focusing on predicate logic, as we will detail in the sections that follow.

In this paper, we are interested in building a model of the student (question asker) in this scenario.
Specifically, we investigate how to enable the student to reason pragmatically about which questions to ask to efficiently acquire knowledge, given only the topic $\mathcal T$ and the conversation history $\mathcal H$.
This setting of \emph{information-seeking conversations} involves many interesting and challenging problems in natural language processing:
\begin{itemize}
\setlength\itemsep{0em}
\item \textbf{Quantifying textual information.} We need to be able to quantify how much knowledge the student has acquired from $\mathcal K$.
\item \textbf{Evaluating language generation when a single reference is insufficient.} At any state in the conversation, there is usually more than one valid question, some more effective and more appropriate than others.
To address this problem, we need to come up with evaluation metrics and objective functions accordingly, rather than relying on the similarity between generated questions and the single reference that is available in existing datasets.
\item \textbf{Pragmatic reasoning with partial information and a large search space.} In order to train computational agents capable of pragmatic reasoning, previous work typically takes the approach of either limiting the space of referents, or the space of possible utterances, or both.
However, the former is infeasible in a communication setting as the student does not have access to $\mathcal K$ beyond what is already revealed in the conversation, and the latter is also impractical for natural conversations that cover a diverse set of topics.
\end{itemize}

We address these challenges by proposing two automatic reward functions that evaluate the informativeness and specificity of questions, and optimizing them with reinforcement learning.

\begin{figure*}
    \centering
    \subfigure[Question
    Generator Model]{
    \includegraphics[width=0.48\textwidth]{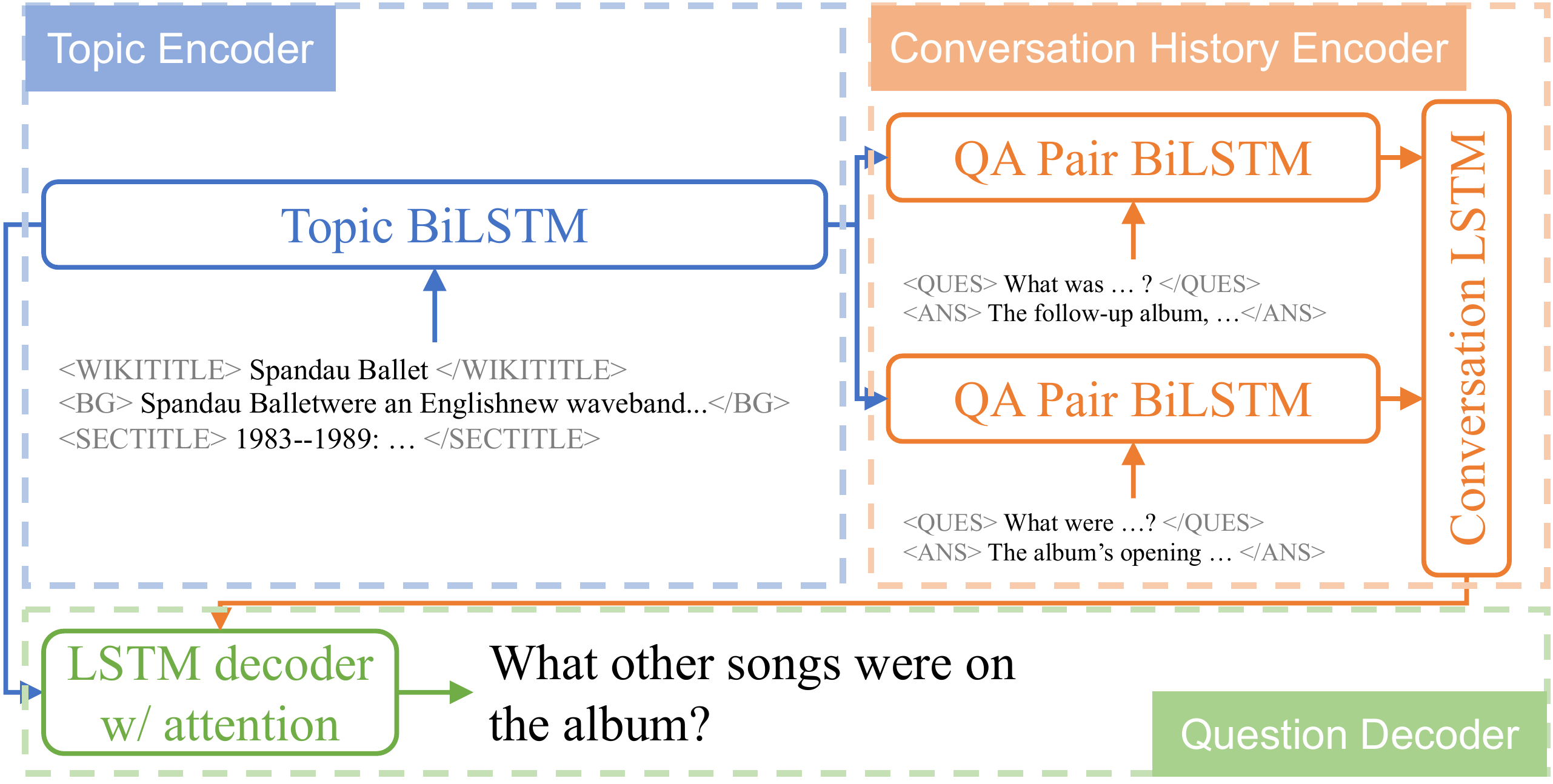}
    \label{fig:architecture:generator}}
    \
    \subfigure[Informativeness and Specificity Model]{
    \includegraphics[width=0.48\textwidth]{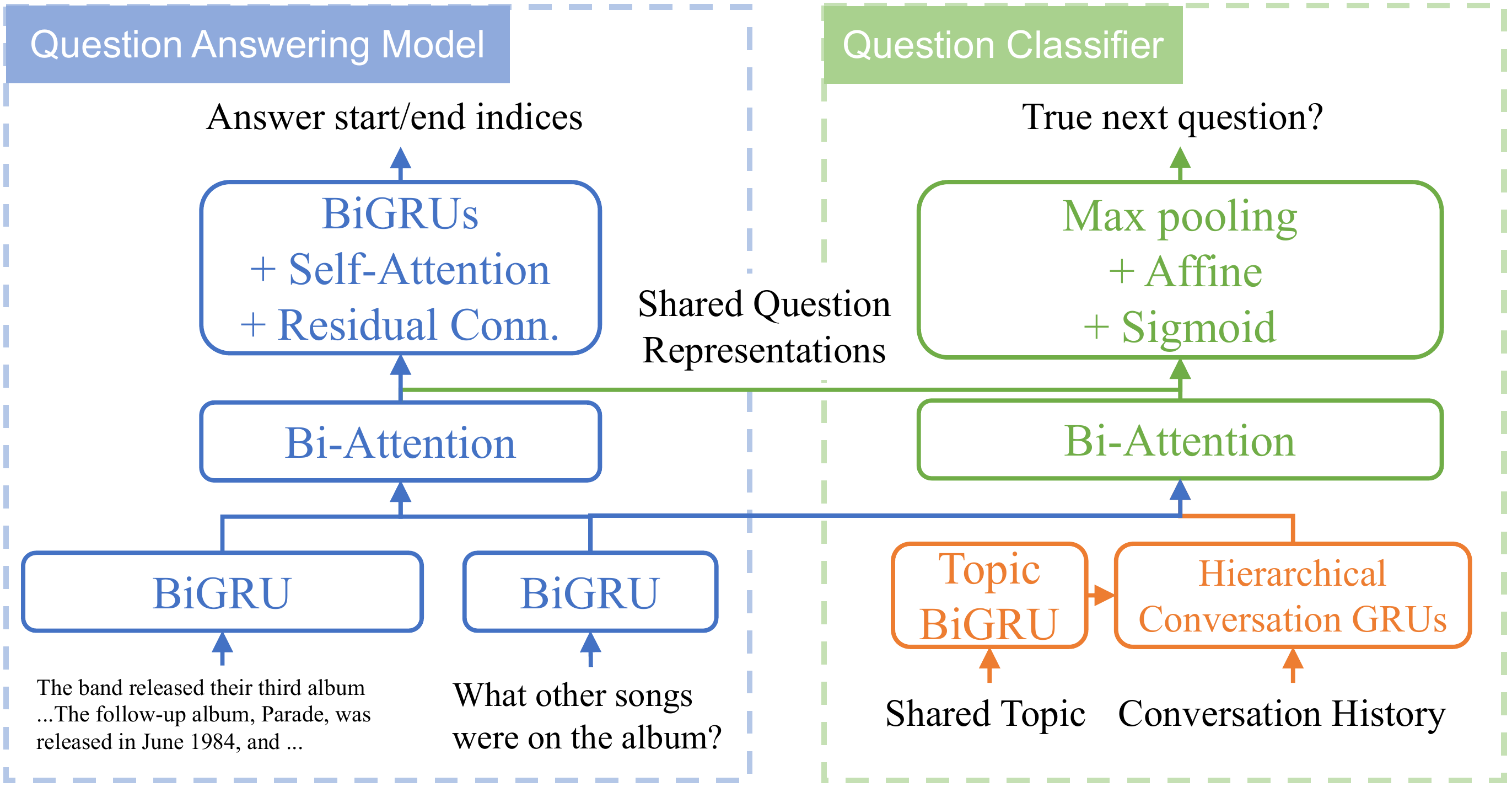}
    \label{fig:architecture:critic}}
    \caption{Model architectures of (a) our question generation model, which takes only the shared topic and conversation history to generate the next question in the conversation; and (b) the model to evaluate how informative and specific generated questions are.}\label{fig:architecture}
\end{figure*}

\subsection{Generating Questions in Conversations}

Before we delve into the proposed approaches for training a question generator model to be pragmatic, an introduction of the model itself is due.

For the purposes of this paper, we assume that the shared topic $\mathcal T$, the shared conversation history $\mathcal H$, and the teacher's knowledge $\mathcal K$ (which the student has no access to) are all made available to agents in natural language.
Since we consider information-seeking conversations only, the conversation history is grouped into pairs of questions and answers: $\mathcal{H} = [(q_1, a_1), (q_2, a_2), \ldots, (q_{|\mathcal H|}, a_{|\mathcal H|})]$.

To generate conversational questions, we build a sequence-to-sequence model that encodes the information available to the student and decodes it into the next question in the conversation (see Figure \ref{fig:architecture:generator}).
Specifically, we first model the shared topic $\mathcal T$ with a bi-directional LSTM (BiLSTM) \cite{hochreiter1997long}, and use the resulting topic representation $h_{\mathcal T}$ in the conversation encoder.
Then we obtain a representation of the conversation with hierarchical LSTM encoders: we first encode each pair of question and answer with $h_{\mathcal T}$ using a BiLSTM, then feed these pair representations into a \emph{unidirectional} LSTM in the direction that the conversation unfolds.
To generate the question, we apply an LSTM decoder with attention both on the topic and the conversation history \cite{bahdanau2015neural}.
This allows us to efficiently batch computation for each conversation by sharing these representations across different turns.
We include detailed description of the model in Appendix \ref{sec:model_details}.

As a baseline, we train this model to minimize the negative log likelihood (NLL) of questions observed in the training set:
\begin{align}
\ell_{\mathrm{NLL}} &= -\frac{1}{N_{\mathrm{p}}}\sum_{i=1}^N\sum_{j=1}^{|\mathcal{H}^{(i)}|} \log P_\theta(q_j^{(i)} | \mathcal{H}^{(i)}_{<j}, \mathcal{T}),
\end{align}
where $\theta$ stands for model parameters, $N$ is the number total conversations in the training dataset, $\mathcal{H}^{(i)}$ the conversation history of the $i$-th conversation in the dataset, and $N_{\mathrm{p}}=\sum_{i=1}^N |\mathcal{H}^{(i)}|$ is the total number of question-answer pairs in the training dataset.
Intuitively, this trains the model to mimic the observed questions in the dataset, but does not provide guarantees or assessment of how well generated questions are actually able to acquire information from the teacher agent.

\subsection{Evaluating Informativeness through Question Answering} \label{sec:informativeness}

In order to train the question generation model to generate pragmatically apt questions that reveal new information from $\mathcal K$, we need to be able to quantify informativeness in communication first.
However, informativeness is difficult to quantify in an open-domain dialogue, and sometimes even subjective.
In this paper, we %
focus on providing an objective metric for how much new information is revealed by a question.
Since questions do not reveal information directly, but rather rely on the answers to them to introduce new facts into the conversation, we begin by defining the \emph{informativeness} of an answer $a$ once it is provided.
Specifically, we are interested in characterizing how much new information an answer $a$ reveals about $\mathcal K$ beyond what is already provided in the conversation history $\mathcal{H}_{<j}$ up until this point in the conversation.
Theoretical quantities like mutual information might seem appealing in this context given their strong grounding in information theory.
However, applying them would potentially require us to fully specify the state space the world can be in for an open-domain conversation, as well as estimating the probability distribution over potential configurations, neither of which is trivial, if feasible.
Therefore, we turn to more practical quantities in defining the informativeness of an answer $a$ given the conversation history $\mathcal{H}_{<j}$ by leveraging the observation that, the more new information an answer reveals about $\mathcal K$, the more likely it involves words that have not already been mentioned in $\mathcal{H}_{<j}$.
Therefore, making use of  the unigram precision function $\mathrm{Prec}(a, a')$ between the predicted answer $a$ and an answer $a'$ that is already provided in the conversation history $\mathcal{H}_{<j}$, we define the informativeness of the predicted answer as follows
\begin{align}
    I_{\mathrm{ans}}(a; \mathcal H_{<j}) &:= 1 - \max_{1\le k<j} \mathrm{Prec}(a, a_k), \label{eqn:informativeness}
\end{align}
Intuitively, the more $a$ overlaps with \emph{any} of the previously revealed answers, the less new information it contains.
This metric of informativeness has the advantages of objectivity and ease of automatic evaluation.
Also note that the choice of unigram precision is here not one of necessity but simplicity and practicality.
It is in principle interchangeable with more sophisticated models of fuzzy text overlap (\eg, BERTScore \citep{zhang2020bertscore}).

We use this definition of answer informativeness to define the utility of potential questions.
Specifically, we define the \emph{informativeness} of a question as the amount of new information it can immediately reveal through its answer
\begin{align}
    I(q; \mathcal{C}_{<j}) &:= I_{\mathrm{ans}}(\mathrm{QA}(q, \mathcal{C}_{< j}), \mathcal{H}_{<j}), \label{eqn:informativeness_final}
\end{align}
where $\mathcal{C}_{<j}=(\mathcal{H}_{< j}, \mathcal{T}, \mathcal{K})$ is the complete context available to the teacher up until the question is raised, $\mathrm{QA}(q, \mathcal{C}_{<j})$ is a pretrained conversational question answering (QA) model that answers the question $q$ from the knowledge source $\mathcal{K}$ given this context.
This is equivalent to using a point estimate for $P(a | q, \mathcal{C}_{<j})$ to evaluate $q$'s expected utility, which is practical for pragmatic reasoning at scale by avoiding the need for aggregating over a large set of candidate answers for each question.
In contrast, previous work on pragmatics often require probabilistic normalization in the space of speaker utterances (questions) and listener actions (answers), which is intractable in our setting.

This definition of informativeness is also \emph{explainable}: it is easy for a human to inspect the answer provided by the QA model and compare it to previous ones to understand how much new information has been revealed.
Note that this definition itself also does not rely on any specific QA model, although more accurate QA models could result in more accurate estimates of informativeness.
For simplicity, we use a bidirectional attention flow model \cite{seo2017bidirectional} with self-attention \cite{clark2018simple} as adapted for conversational QA by \citet{choi2018quac} (see Figure \ref{fig:architecture:critic}).

\subsection{Evaluating Question Specificity}

Now that we have a metric to evaluate informativess, can we maximize it and obtain a good model for generating pragmatic conversational questions?
It turns out that there are two issues with na\"ively optimizing this value: generated questions could be overly generic or disruptive of the conversation flow while still acquiring new information.
For instance, questions like \emph{What else?}\ almost always reveal new information.
On the other hand, in our example in Figure \ref{fig:example}, \emph{Did they go on tour for their 1983 album?}\ seems more disruptive (topic-changing) as the next question in the conversation than the candidate questions in the figure.

To address this, we take a similar approach to previous work by selecting negative examples to target these issues and training a classifier to distinguish them from questions that were actually part of the conversation \cite{lowe2017towards,rao2018learning}.
Once this classifier is trained, we can make use of the score it assigns different candidate questions to evaluate how specific each is to the current conversation history.
Specifically, we select two kinds of negative questions: frequent questions from the training set (frequency$>$1) and random questions other than the observed one from the same conversation.
We train a model (with shared parameters with the QA model, see Figure \ref{fig:architecture:critic}) to assign a probability that a question is the true next question (positive) given the conversation history, and define this quantity as the \emph{specificity} of the question
\begin{align}
S(q; \mathcal{H}_{<j}, \mathcal{T}) &:= P_\xi(q\textrm{ is positive}|\mathcal{H}_{<j}, \mathcal{T}),
\end{align}
where $\xi$ is the parameters of the classifier optimized with binary cross entropy loss.
Once this classifier is trained jointly with the QA model, we can use this specificity reward to bias the model towards generating questions that are not only informative, but also specific to the given conversation history.

\diff{}{
Conceptually, our specificity idea is related to a few separate but connected concepts in NLP, namely discourse coherence, relevance, and reducing genericness in natural language generation.

The coherence and relevance of a piece of text in a discourse is highly correlated with the perceived quality of the generated text.
Previous work has approached generating coherent utterances in conversations through encouraging the model to learn similar distributed representations throughout the conversation \citep{baheti-etal-2018-generating, xu-etal-2018-better, zhang2018reinforcing}.
In contrast, we achieve the same goal with a discriminative classifier, which is trained to contrast the true follow-up question (relevant and coherent) against randomly sampled questions (irrelevant) from other conversations and out-of-order questions (uncoherent).
The idea of discerning discourse consistency has also been applied to large pretrained language models \citep{devlin2019bert,iter2020pretraining}, which is demonstrated to sometimes yield performance gains when the they are finetuned on downstream tasks.

On the other hand, since we sample frequent questions in the training set as negative examples for the classifier, it also discourages the model from generating overly generic questions.
Previous work has attacked the problem of  genericness in conversational natural language generation by proposing auxiliary training objectives, \eg, ones that maximize the utility of the generated utterance estimated with adversarial networks \citep{rao-daume-iii-2019-answer}, specificity estimates that are estimated from data \citep{ko2019domain, ko2019linguistically}, or the mutual information between the generated turn and previous ones \citep{li2016deep}.
Our proposed method can be viewed as a generalization of these approaches, where the objective to be optimized at the time of generation is implicitly specified via a parameterized model by choosing negative examples for contrast.
}

\subsection{Generating Informative and Specific Questions}

Given the informativeness metric and specificity reward, we can improve upon these by %
maximizing the following reward function that blends the two in a weighted sum
\begin{align}
    R(q; \mathcal{C}_{<j}) &= \lambda_1 I(q; \mathcal{C}_{<j}) + (1-\lambda_1) S(q; \mathcal{H}_{<j}, \mathcal{T}).
\end{align}
Since this quantity can only be evaluated once a complete question has been generated, the non-differentiability of the decoding process prevents us from directly optimizing it with respect to $\theta$ using gradient-based optimization.
However, we can still estimate the gradient of the expected reward of generated questions, $\mathbb{E}_{q\sim P_\theta}[R(q)]$ using REINFORCE \cite{williams1992simple}, a reinforcement learning technique.
For an example $q$, the gradient estimate is the gradient of the following loss function
\begin{align}
\ell_{{R}}(q) &= -\big[{R}(\hat{q})-b(q)\big]\log P_\theta(\hat{q})
\end{align}
where $\hat{q}$ is a sample from $P_\theta$ and we dropped the dependency on $\mathcal{C}_{<j}$ for notational clarity.
$b(q)$ is called the \emph{baseline function}, which, if chosen carefully, reduces the variance of this gradient estimate and results in faster convergence.
We apply a technique called self-critical sequence training \cite{rennie2017self}, which selects $b(q)={R}(q_{\mathrm{G}})$, the reward obtained by the greedily decoded sequence, $q_G$, from the question generator.

To ensure that the generator maximizes the desired reward function without losing fluency in generated questions, we combine $\ell_{{R}}$ with negative log likelihood during model finetuning \cite{paulus2017deep}.
We finetune a pretrained question generator (with $\ell_{\mathrm{NLL}}$) using the following objective
\begin{align}
\ell &= \lambda_2 \ell_{{R}} + (1-\lambda_2) \ell_{\mathrm{NLL}}.
\end{align}
Here, $\ell_{\mathcal{R}} =\frac{1}{N_{\mathrm{p}}}\sum_{i=1}^N\sum_{j=1}^{M_i} \ell_{\mathcal{R}}(q_j^{(i)})$. We choose $\lambda_1 = 0.5$ and $\lambda_2 = 0.98$ in our experiments, which were chosen by tuning the model on the dev set.

\section{Experiments}
\label{sec:experiments}

\paragraph*{Data.}
For our experiments, we use the \quac{} dataset presented by \citet{choi2018quac}.
Although other similar datasets share some common characteristics, some crucial differences render them inapplicable for our experiments.
For instance, \coqa{} \cite{reddy2019coqa} gives both agents access to the context, while
Wizard of Wikipedia \cite{dinan2018wizard} does not assign the student agent clear goals of acquiring new information.

Since \quac{}'s test set is held private for fair evaluation, for this work we repurpose the original dev set as our test set.
We randomly split the training set into  training and development partitions, ensuring that the Wikipedia entities discussed in conversations do not overlap between these partitions.
The goal of the split is to obtain a development set that is roughly as large as the repurposed test set.
The statistics of our data split can be found in Table \ref{tab:quac_split}.

\begin{table}
    \centering \small
    \begin{tabular}{llccc}
    \toprule
    Split & Orig. & \# Entities & \# Dialogues & \# QA pairs \\
    \midrule
    Train & Train & 2727 & 10572 & 76338 \\
    Dev & Train & \phantom{0}264 & \phantom{00}995 & \phantom{0}7230 \\
    Test & Dev & \phantom{0}721 & \phantom{0}1000 & \phantom{0}7354 \\
    \bottomrule
    \end{tabular}
    \caption{Data split of the \quac{} dataset used in our experiments.} \label{tab:quac_split}
\end{table}

\paragraph*{Training.}
We follow the recipe available in AllenNLP \cite{gardner2017allennlp} to train the QA model on \quac{}, and make sure that it obtains performance on par with that reported by \citet{choi2018quac} on the official dev set (with multiple answer references).%
\footnote{However, in practice, we remove the ELMo component, which greatly speeds up computation at the cost of only losing 2--3 \fone{} in answer prediction.}
We use the Adam optimizer \cite{kingma2015adam} with default hyperparameters to train and finetune our question generator, and anneal the learning rate by 0.5 whenever dev performance does not improve for more than 3 consecutive epochs (patience=3).
When training finishes, the specificity classifier achieves approximately 75\% \fone{} on the dev set when the true next question, sampled frequent questions and random questions from the same conversation have a balanced ratio of 1:1:1.
For unanswerable questions in \quac{}, we revise Equation (\ref{eqn:informativeness_final}) and set informativeness to zero if the predicted answer is \texttt{CANNOTANSWER}, as the answer does not reveal new information about the hidden knowledge $\mathcal K$.

\section{Results}
\label{sec:results}

\begin{table*}[t]
    \centering
    \begin{tabular}{lccccccccc}
        \toprule
        \multirow{2}{1.8cm}{System} & \multicolumn{4}{c}{dev} & \phantom{.} & \multicolumn{4}{c}{test} \\
        \cline{2-5}\cline{7-10}
        & \textsc{pplx} & \textsc{rouge-l} & \textsc{info} & \textsc{spec} && \textsc{pplx} & \textsc{rouge-l} & \textsc{info} & \textsc{spec} \\
        \midrule
        Reference & -- & -- & 0.639 & 0.825 && -- & -- & 0.635 & 0.821\\
        \midrule
        Baseline & \textbf{8.46} & 25.5 & 0.653 & 0.761 && \textbf{8.08} & 24.3 & 0.664 & 0.779 \\
        Our Model & 9.24 & \textbf{25.6} & \textbf{0.740} & \textbf{0.834} && 8.79 & \textbf{24.9} & \textbf{0.752} & \textbf{0.835} \\
        \bottomrule
    \end{tabular}
    \caption{Evaluation of the baseline system, our pragmatically finetuned system, and the reference questions on conventional metrics as well as our proposed metric and reward functions.}
    \label{tab:main_results}
\end{table*}

\subsection{Metric-based Evaluation}

For the baseline model and our model finetuned for informativeness and specificity, we generate predictions with greedy decoding for simplicity.
We evaluate them %
on conventionally used metrics such as perplexity (\textsc{pplx}) of the reference question and the \fone{} score of the ROUGE-L metric (\textsc{rouge-l}) \cite{lin2004rouge} between the predicted questions and the reference.
The former helps verify the overall quality of our model, while the latter helps us compare single-reference metrics to our proposed ones.
We also report the informativeness metric (\textsc{info}) and specificity reward (\textsc{spec}) for these models, and compare them to the reference questions on these measures on both the dev and test sets.

As shown in Table \ref{tab:main_results}, the baseline model and our pragmatically finetuned model achieve comparable performance when evaluated against the reference question using $n$-gram overlap metrics (\textsc{rouge-l}), and the perplexity of the reference question is only slightly worse.
As expected, these metrics tell us nothing about how well the model is going to fare in actual communication, because perplexity does not evaluate the usefulness of generated questions, and \textsc{rouge-l} can barely tell these systems apart.

We can also see in Table \ref{tab:main_results} that the finetuned model improves upon the baseline model on both  informativeness and specificity.
Further, we notice that despite their high specificity, the reference questions are only about as informative as our baseline questions on average, which is a bit surprising at first sight.
Further analysis reveals that about 12.6\% of dev questions and 15.7\% test ones are considered unanswerable by crowd workers, which is a byproduct of the information-asymmetric setting adopted when the data was collected.
As a result, many reference questions could be considered uninformative by our definition, since they might cause the QA model to abstain from answering.

\subsection{Human Evaluation}

\begin{table}
    \setlength{\tabcolsep}{10pt}
    \centering
    \small
    \diff{
    \begin{tabular}{lc<{\hspace{-7pt}}c<{\hspace{-7pt}}c<{\hspace{-7pt}}p{1.1cm}<{\hspace{-10pt}}}
    \toprule
    & Win & Tie & Lose & $p$-value \\
    \midrule
    \multicolumn{5}{c}{Ours vs. Baseline}\\
    \midrule
    Overall  & 30.0\% & 48.0\% & 22.0\% & 0.108\\
    \textsc{info}  & 21.0\% & 68.5\% & 10.5\% & 0.015 \\
    \textsc{spec} & 26.0\% & 53.5\% & 20.5\% & 0.171 \\
    \midrule
    \multicolumn{5}{c}{Ours vs. Human Reference}\\
    \midrule
    Overall & 16.0\% & 25.0\% & 59.0\% & $<10^{-6}$ \\
    \textsc{info} & \phantom{0}9.0\% & 72.0\% & 19.0\% & 0.011 \\
    \textsc{spec} & 13.0\% & 27.5\% & 59.5\% & $<10^{-6}$ \\
    \bottomrule
    \end{tabular}\\}
    {
    \resizebox{0.48\textwidth}{!}{%
    \begin{tabular}{lc<{\hspace{-7pt}}c<{\hspace{-7pt}}c<{\hspace{-7pt}}p{1.2cm}<{\hspace{-10pt}}}
    \toprule
    & Win & Tie & Lose & $p$-value \\
    \midrule
    \multicolumn{5}{c}{Ours vs. Baseline}\\
    \midrule
    Overall  & 29.00\% & 52.50\% & 18.50\% & 0.006\\
    \textsc{info}  & 19.25\% & 74.50\% & \phantom{0}6.25\% & 3$\times$10\textsuperscript{-5}\\
    \textsc{spec} & 25.50\% & 56.50\% & 18.00\% & $0.029$\\
    \midrule
    \multicolumn{5}{c}{Ours vs. Human Reference}\\
    \midrule
    Overall & 17.75\% & 26.25\% & 56.00\% &$<$2$\times$10\textsuperscript{-6} \\
    \textsc{info} & \phantom{0}8.75\% & 74.50\% & 19.75\% & 3$\times$10\textsuperscript{-4} \\
    \textsc{spec} & 15.25\% & 29.00\% & 55.75\% & $<$2$\times$10\textsuperscript{-6} \\
    \bottomrule
    \end{tabular}
    }
    }
    \caption{Human evaluation comparing questions our system generated to those from the baseline, as well as the original reference questions in \quac{}. We perform a bootstrap test with $10^6$ samples for the difference between pairs of systems and report the $p$-values here.} \label{tab:human_eval}
\end{table}

Although the results in Table~\ref{tab:main_results} show that our model sees substantial improvements on the proposed informativeness and specificity metrics, it remains unclear whether these improvements correlate well with human judgement of quality, which is critical in the application of the resulting system.
To study this, we conduct a comparative human evaluation.

We randomly selected 200 turns from the test set, and asked two NLP PhD students to evaluate the reference questions, as well as those generated by the baseline model and our model.
These questions are evaluated on their overall quality, informativeness, and specificity, where the annotators are asked to rank the candidate questions on each metric with ties allowed.
System identity is hidden from the annotators, and the order of the systems is shuffled for each comparison.
Prior to annotation, both annotators were educated to follow the same guidelines to encourage high agreement (see Appendix \ref{sec:human_eval} for details).

As shown in Table \ref{tab:human_eval}, human annotators favor our system over the baseline on informativeness (\diff{89.5\%}{93.75\%} of our questions are considered equally or more informative), and to a lesser extent, overall quality (\diff{78.0\%}{81.5\%}) and specificity (\diff{79.5\%}{82\%}).
\diff{}{We find that 26.1\% of questions generated by these systems are identical, which inflates the number of ties in human evaluation.
We expect a starker contrast if a sampling-based decoding strategy were applied for generation diversity, which we leave to future work.}
\diff{This difference is partly due}{We also attirbute this difference in human-perceived quality on these three aspects partly} to the inherent nature of these annotation tasks: while our annotators agree on \diff{77.3\%}{80.3\%} of the pair-wise judgments regarding informativeness, agreement decreases to \diff{71.7\%}{70.7\%} for overall quality and \diff{70.3\%}{69.2\%} for specificity since they are more subjective.
It is encouraging, however, that our system is also considered equally or more informative than the human reference \diff{81\%}{80.25\%} of the time.
What negatively affects human's perception of the overall quality of questions our system generates is largely attributable to the over-genericness of these questions compared to the references\diff{.}{, and a sometimes blatant lack of common sense (\eg, questions like \emph{``What did he do after his death?''}).}

\section{Analysis}
\label{sec:analysis}

\begin{figure}[!t]
    \centering
    \resizebox{0.49\textwidth}{!}{%
    \footnotesize
    \begin{tabular}{|p{7.7cm}|}
        \hline
        \textbf{Background:} Spandau Ballet (English band)\\
        \textbf{Topic:} 1983--1989: International success and decline\\
        \hline
        Candidate Questions\\
        \textbf{BL\textsubscript{1}:} {What happened in 1983?}\\
        \textbf{Ours\textsubscript{1}:} {What happened in 1983?}\\
        \textbf{Ref\textsubscript{1}:} What was the first indication of Spandau Ballet's success at the international level?\\
        \hline
        \textbf{Ans\textsubscript{1}:} The follow-up album, \emph{Parade}, was released in June 1984, and its singles were again big successes in the charts in Europe, Oceania and Canada.
        \\
        \hline
        \textbf{BL\textsubscript{2}:} {What was the name of the album?}\\
        \textbf{Ours\textsubscript{2}:} What was the most popular single from the album?\\
        \textbf{Ref\textsubscript{2}:}  What were the notable songs from the album Parade?\\
        \hline
        \textbf{Ans\textsubscript{2}:} The album's opening song, \emph{``Only When You Leave''}. \\
        \hline
        \textbf{BL\textsubscript{3}:} {What was the name of the album that was released?}\\
        \textbf{Ours\textsubscript{3}:} What other songs were on the album?\\
        \textbf{Ref\textsubscript{3}:} How did the opening song do on the charts? \\
        \hline
        \textbf{Ans\textsubscript{3}:} Became the band's last American hit.\\
        \hline
        \textbf{BL\textsubscript{4}:} What was the last album that they released?\\
        \textbf{Ours\textsubscript{4}:} What other songs were on the album?\\
        \textbf{Ref\textsubscript{4}:} {Are there any other interesting aspects about this article?}\\
        \hline
    \end{tabular}
    }
    \caption{A success example where our automatic metrics align well with human judgement of informativeness and specificity, when comparing questions generated by the baseline (BL), our system (Ours), and the original human-written reference (Ref).
    \diff{}{All answers in the figure are from the original human-human conversations in \quac{} in answer to reference questions.}}
    \label{fig:success}
\end{figure}

We further analyze concrete examples of generated questions in conversations to understand the behavior of our informativeness and specificity metrics.

\paragraph{Case Study.}
To sanity check whether our informativeness metric and specificity reward match human intuition, we manually inspect a few examples from the test set.
Figure \ref{fig:success} represents a case where all the questions our system generated are considered equal to or more informative than the reference and baseline generated questions by our metric.
As shown in the example, the baseline system is prone to generating topical but uninformative questions (BL\textsubscript{2} and BL\textsubscript{3}).
Our system finetuned on our reward function is more pragmatic and asks about relevant questions that can likely be answered from the unseen paragraph $\mathcal K$.
Our informativeness metric also correctly identifies that both Ours\textsubscript{3} and Ref\textsubscript{3} are good questions that reveal new information about $\mathcal K$, although there is very little overlap between the two.
On the other hand, the specificity reward successfully identifies that BL\textsubscript{3} and Ref\textsubscript{4} are the least specific questions of their respective turn, where the former is disconnected from the most recent topic under discussion (the song), the latter is phrased in an overly generic way.

We also demonstrate some clear failure cases.
In Figure \ref{fig:failure}, we see that our informativeness and specificity measures make judgements a human will unlikely make, as the topic implies $\mathcal K$ is unlikely to contain information about Moyet's first album/recording.
In fact, the QA model fails to recognize that these questions (BL\textsubscript{1,2}, Ours\textsubscript{1,2,3}, Ref\textsubscript{1}) are unanswerable, and instead assigns them high informativeness.
The specificity model, on the other hand, fails to recognize near paraphrases (BL\textsubscript{1} vs Ours\textsubscript{1}) and a question that was likely just answered (BL\textsubscript{3}).
A positive finding in this example is that the informativeness metric is well-aligned with pragmatic behavior in the fourth turn---had Moyet won the Grammy, the previous answer (A\textsubscript{3}) would have mentioned it instead of just her nomination.

We include the answering contexts for these examples in Appendix \ref{sec:answering_contexts} for the reader's reference.

\begin{figure}[!t]
    \centering
    \resizebox{0.49\textwidth}{!}{%
    \footnotesize
    \begin{tabular}{|p{7.7cm}|}
        \hline
        \textbf{Background:} Alison Moyet (English singer)\\
        \textbf{Topic:} 1990s: Further recordings and hiatus\\
        \hline
        Candidate Questions\\
        \textbf{BL\textsubscript{1}:} What was the first album released?\\
        \textbf{Ours\textsubscript{1}:} What was her first recording?\\
        \textbf{Ref\textsubscript{1}:} What did she do in 1990?\\
        \textit{\textsc{info}: Ref $=$ BL $=$ Ours \qquad \textsc{spec}: Ref $\approx$ Ours $>$ BL}\\
        \hline
        \textbf{Ans\textsubscript{1}:} \texttt{CANNOTANSWER}
        \\
        \hline
        \textbf{BL\textsubscript{2}:} What was her first album?\\
        \textbf{Ours\textsubscript{2}:} What was her first album?\\
        \textbf{Ref\textsubscript{2}:} What else did she record in the 1990's?\\
        \textit{\textsc{info}: Ref $=$ BL $=$ Ours  \qquad \textsc{spec}: BL$=$Ours$>$Ref }\\
        \hline
        \textbf{Ans\textsubscript{2}:} Hoodoo.\\
        \hline
        \textbf{BL\textsubscript{3}:} What was the name of the album?\\
        \textbf{Ours\textsubscript{3}:} What was her first album? \\
        \textbf{Ref\textsubscript{3}:} Did he reach bill board charts?\\
        \textit{\textsc{info}: BL $=$ Ref $>$ Ours  \qquad \textsc{spec}: Ours$>$BL$>$Ref }\\
        \hline
        \textbf{Ans\textsubscript{3}:} Yes, and Moyet was nominated for a Grammy for the single\\
        \hline
        \textbf{BL\textsubscript{4}:} What was the song called?\\
        \textbf{Ours\textsubscript{4}:} What other songs did she release?\\
        \textbf{Ref\textsubscript{4}:} Did she receive an award? \\
        \textit{\textsc{info}: BL $=$ Ours $>$ Ref  \qquad \textsc{spec}: BL$\approx$Ours$\approx$Ref }\\
        \hline
    \end{tabular}
    }
    \caption{A failure case of our informativeness metric and specificity reward, when comparing reference questions (Ref), baseline generated questions (BL), our those generated by our model (Ours).
    \diff{}{All answers in the figure are from the original human-human conversations in \quac{} in answer to reference questions.}}
    \label{fig:failure}
\end{figure}

\paragraph{Explainable Informativeness.}
As stated in Section \ref{sec:informativeness}, our definition of informativeness is explainable to humans---we demonstrate this with concrete examples.
For instance, in the example in Figure \ref{fig:success}, although the question \emph{What happened in 1983?}\ is phrased rather vaguely, the QA model is able to identify its correct answer from the paragraph \emph{The band released their third album, True, in March 1983}, which offers new information \diff{}{(note the answer in the figure only reflects the actual human-human conversation, not this hypothetical one)}.
Similarly, the QA model correctly identifies that the question our model generated on the second turn (Ours\textsubscript{2}) has the same answer as the human reference (Ref\textsubscript{2}), which introduces a new entity into the conversation.
BL\textsubscript{2} and BL\textsubscript{3} are deemed uninformative in this case since the QA model offered the same answer about the album \emph{True} again.
Although this answer is about an incorrect entity in this context (the album \emph{True} instead \emph{Parade}, which is the focus of discussion), the large amount of overlap between this answer and Ans\textsubscript{1} is still sufficient to regard these questions as less informative.

We note that this informativeness metric does have an exploitable flaw---it does not prevent the questioner from asking vague, open-ended questions (\eg, \emph{What else do you know?}) to acquire knowledge.
In fact, we find this strategy is also adopted by \quac{}'s crowd workers.
However, our specificity reward penalizes genericness, and therefore alleviates this issue in the questions our system generates.
We show that our system repeats $n$-grams from previous questions less frequently, and refer the reader to Appendix \ref{sec:specificity} for details.

\section{Conclusion}

In this paper, we presented a question generation system in information-seeking conversations.
By optimizing our proposed automatic metrics for informativeness and specificity, the model is able to generate pragmatically relevant and specific questions to acquire new information about an unseen source of textual knowledge.
Our proposed method presents a practical if shallow implementation of pragmatics in an open-domain communication setting beyond simple reference games.
We hope that our work brings the community's attention to this important problem of natural language communication under information asymmetry.

\diff{}{
  \section*{Acknowledgments}
The authors would like to thank Christopher Potts, Kawin Ethayarajh, Alex Tamkin, Allen Nie, and Emrah Budur, among other members of the Stanford NLP Group, for their feedback on earlier versions of this paper.
We are also grateful to the anonymous reviewers for their insightful comments and constructive discussions.
This research is funded in part by Samsung Electronics Co., Ltd.\ and in part by the SAIL-JD Research Initiative.
}

\bibliography{qugc-paper}
\bibliographystyle{acl_natbib}

\clearpage
\appendix
\section{Model Details} \label{sec:model_details}

In this section, we include the details of the question generation model and the informativeness/specificity model we used in our experiments.

\subsection{Question Generator}

For the input to the encoder and decoder models in our question generator, we tokenize them with the spaCy toolkit,\footnote{\url{https://spacy.io/}} and initialize word representations with 100-dimensional GloVe vectors \cite{pennington2014glove}.
As shown in Figure \ref{fig:architecture}, we also introduce special XML-like symbols to delimit different parts of the input to various models.
The representations of these special symbols are randomly initialized, and finetuned with those of the top 1000 most frequent words in the training set during training.

For the topic containing the title of the Wikipedia page and the background on the entity after concatenating them with special symbols, we feed them into a topic BiLSTM model and obtain the topic representation with a multi-layer perceptron (MLP) attention mechanism, using the concatenated final state from each direction of the BiLSTM as the key

\begin{align}
h_{\mathcal{T}} & = \mathrm{BiLSTM}_{\mathcal{T}}(x_{\mathcal{T}}),\\
h_{\mathcal{T}}^{\mathrm{attn}} &= \mathrm{MLPSelfAttn}(h_{\mathcal{T}}).
\end{align}

We use this representation to initialize the BiLSTM we use to encode each pair of turns in the conversation that contains a question and its corresponding answer
\begin{align}
h_{\mathcal{H}_j}^0 &= \mathrm{BiLSTM}_{\mathcal{H},\mathrm{pair}}(x_{\mathcal{H}_j}, h_{\mathcal{T}}^{\mathrm{attn}}),
\end{align}
which we in turn use as the input to our unidirectional LSTM model to obtain the representation of the entire conversation up until a certain turn
\begin{align}
h_{\mathcal{H}} &= \mathrm{BiLSTM}_{\mathcal{H}, \mathrm{conv}}([h_{\mathcal{H}_1}^0, \cdots, h_{\mathcal{H}_{|\mathcal{H}|}}^0]).
\end{align}
Note that for modeling simplicity, we use the section title as the 0\textsuperscript{th} ``turn'' for each conversation.
We similarly obtain a summary representation of the conversation with MLP self-attention
\begin{align}
h_{\mathcal{H}}^{\mathrm{attn}} &= \mathrm{MLPSelfAttn}(h_{\mathcal{H}}),
\end{align}
and concatenate it with $h_{\mathcal{T}}^{\mathrm{attn}}$ to initialize the decoder.

To represent the input words in the decoder, we use the same embedding matrix as the encoder.
We also employ weight tying between the input embeddings and the output weights for word prediction to reduce parameter budget \cite{press2017using}.
For each word in the decoder input, we concatenate its embedding with $h_{\mathcal{T}}^{\mathrm{attn}}$ for topical context.
We provide the decoder access through attention to all of the representations of encoded tokens, \ie, $[h_{\mathcal{T}}^{\mathrm{attn}}, h_{\mathcal{H}_0}^0, \cdots, h_{\mathcal{H}_{|\mathcal{H}|}}^0]$.
Finally, the weighted average of encoder representations is combined with the decoder LSTM's representation of the decoded sequence to yield a probabilistic distribution over words in the vocabulary.

\subsection{Informativeness/Specificity Model}

For informativeness, we follow closely the open implementation of BiDAF++ for \quac{} that is available in AllenNLP \cite{gardner2017allennlp}.
For each word, we concatenate its word representations with character representations derived from a convolutional neural network from its character spelling.
We replace the ELMo embeddings with GloVe ones for computational efficiency, which results in a relatively small drop in QA performance compared to AllenNLP's implementation (by about 2--3 \fone on the official dev set).
Note that following \citet{choi2018quac}, we use gated recurrent units \cite[GRUs; ][]{cho2014learning} in this part of the model.

For the specificity model, we first encode the topic and conversation history in a similar fashion as we did for the encoder in the question generator.
Then, this representation is combined with the question representation from the BiGRU encoder in the QA model via a bidirectional attention (bi-attention) mechanism.
The resulting representation is combined with the question representation from the bi-attention in the QA model, and max pooled over time, before an affine transform is applied to convert the representation into a score.

\section{Contexts for Case Study Examples} \label{sec:answering_contexts}

We include in Figures \ref{fig:success_context} and \ref{fig:failure_context} the contexts that contain the answer for the examples we studied in Section \ref{sec:analysis}, with gold answers in the case study highlighted in the paragraphs.
Following \citet{choi2018quac}, we concatenate an artificial \texttt{CANNOTANSWER} token to the end of the paragraph for the question answering model to abstain from answering the question.

\begin{figure}[!t]
    \centering
    \footnotesize
    \begin{tabular}{|p{7.2cm}|}
        \hline
        The band released their third album, True, in March 1983. Produced by Tony Swain and Steve Jolley, the album featured a slicker pop sound. It was at this point that Steve Norman began playing saxophone for the band. Preceded by the title track which reached number one in various countries, the album also reached number one in the UK. Their next single, ``Gold'', reached number 2. \answer{1}{The follow-up album, Parade, was released in June 1984, and its singles were again big successes in the charts in Europe, Oceania and Canada.} \answer{2}{The album's opening song, ``Only When You Leave''}, \answer{3}{became the band's last American hit.} At the end of 1984, the band performed on the Band Aid charity single and in 1985 performed at Wembley Stadium as part of Live Aid. During this same year, Spandau Ballet achieved platinum status with the compilation The Singles Collection, which kept the focus on the band between studio albums and celebrated its five years of success. However, the album was released by Chrysalis Records without the band's approval and the band instigated legal action against the label. In 1986, Spandau Ballet signed to CBS Records and released the album Through the Barricades, in which the band moved away from the pop and soul influences of True and Parade and more toward rock. Though the first single, ``Fight for Ourselves'' peaked at 15 in the UK, the title track and the album both reached the Top 10 in the UK and Europe. After a hiatus from recording, the band released their next album, Heart Like a Sky, in September 1989. The album and its singles were unsuccessful in the UK, and the album itself was not released in the United States. It did, however, do well in Italy (where its singles ``Raw'' and ``Be Free with Your Love'' reached the Top 10) and also in Belgium, Germany and the Netherlands. CANNOTANSWER\\
        \hline
    \end{tabular}
    \caption{Private context that contains the answers to questions in our case study example in Figure \ref{fig:success}.}
    \label{fig:success_context}
\end{figure}

\begin{figure}[!t]
    \centering
    \footnotesize
    \begin{tabular}{|p{7.2cm}|}
        \hline
        Following a period of personal and career evaluation, \answer{2}{Hoodoo} was released in 1991. The album sold respectably in the UK, \answer{3}{and Moyet was nominated for a Grammy for the single ``It Won't Be Long''.} However, the release of Hoodoo marked the beginning of an eight-year fight for Moyet to secure complete control of her artistic direction. Like many similar artists (including Aimee Mann and the late Kirsty MacColl), Moyet was reluctant to record a radio-friendly ``pop'' album simply for the sake of creating chart hits. Moyet's next album, Essex (1994), was also a source of controversy for her; in order for the album to be released, her label(now Sony) insisted that certain Essex tracks be re-recorded and re-produced, and that there be additional material remixed to create a more' commercial' package. The video for the single ``Whispering Your Name'' again featured Dawn French. Following the release of Essex, Sony released a greatest hits compilation of Moyet's work. Singles entered the UK charts at No. 1 and, following a UK tour, was re-issued as a double CD set which included ``Live (No Overdubs)'', a bonus live CD. Upon re-issue, Singles charted again, this time in the Top 20. Due to prolonged litigation with Sony, Moyet did not record or release a new studio album for over eight years after the release of Essex. During this time, however, she recorded vocals for Tricky, Sylk-130, Ocean Colour Scene, The Lightning Seeds, and King Britt, and was featured on the British leg of the Lilith Fair tour. 2001 saw the release of The Essential Alison Moyet CD, and in 2002 The Essential Alison Moyet DVD. In 1995, she sang back-up vocals with Sinead O'Connor for one of Dusty Springfield's last television appearances, singing ``Where Is a Woman to Go ?'' on the music show Later With Jools Holland. \answer{1}{CANNOTANSWER}\\
        \hline
    \end{tabular}
    \caption{Private context that contains the answers to questions in our case study example in Figure \ref{fig:failure}.}
    \label{fig:failure_context}
\end{figure}

\section{Specificity Analysis} \label{sec:specificity}

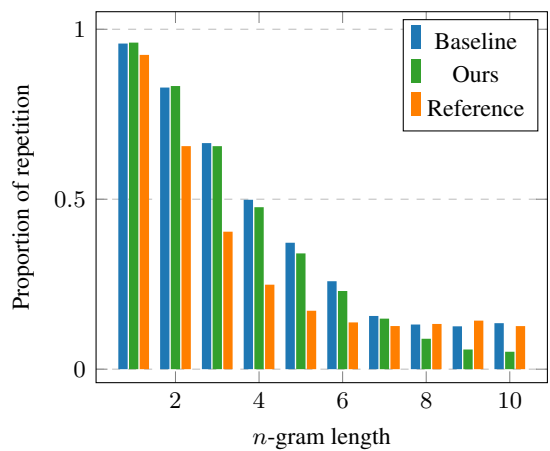
\begin{figure}[!h]
    \pgfplotstableread[row sep=\\,col sep=&]{
	x & Ref & BL & Ours \\
	1 & 0.9245430146525461 & 0.9579253823605656 & 0.960670558737944 \\
2 & 0.6561414521866499 & 0.828484636055614 & 0.8329562036077569 \\
3 & 0.4044315849345896 & 0.6650489546006039 & 0.6560488818771442 \\
4 & 0.24854198773154443 & 0.4983444298073842 & 0.4764725743577151 \\
5 & 0.17187923884819595 & 0.37192593306972704 & 0.34056046613024693 \\
6 & 0.13754508805431784 & 0.25869628763519437 & 0.22994237588652483 \\
7 & 0.12685696507738975 & 0.15651946694538804 & 0.14865331491712708 \\
8 & 0.13301029678982434 & 0.1313503553566728 & 0.08945488429880416 \\
9 & 0.14279983955074207 & 0.12605042016806722 & 0.057837660203746304 \\
10 & 0.12671480144404332 & 0.13529411764705881 & 0.05124340617935192 \\
}\mydata

\pgfplotsset{compat=1.11,
	/pgfplots/ybar legend/.style={
		/pgfplots/legend image code/.code={%
			\draw[##1,/tikz/.cd,bar width=3.5pt,yshift=-0.2em,bar shift=0pt]
			plot coordinates {(0cm,0.8em)};},
	},
}

\begin{tikzpicture}[font=\small]
\begin{axis}[
ybar=.5pt,
xlabel={$n$-gram length},
ylabel={Proportion of repetition},
height=6.5cm,
/pgf/bar width=3.5pt,
xtick align=inside,
ymajorgrids=true,
grid style=dashed,
]

\addplot[draw opacity=0, fill=blue] table[x=x,y=BL]{\mydata};
\addplot[draw opacity=0, fill=green] table[x=x,y=Ours]{\mydata};
\addplot[draw opacity=0, fill=orange] table[x=x,y=Ref]{\mydata};
\legend{Baseline, Ours, Reference}

\end{axis}
\end{tikzpicture}
    \caption{Proportion of repeated $n$-grams in questions from the conversation history.
    As can be seen from the plot, our pragmatic system reduces the amount of $n$-grams repeated from previous questions especially for longer $n$-grams.} \label{fig:ngram_overlap}
\end{figure}

We examine the outputs of our model to assess whether finetuning on the specificity reward results in more specific questions rather than generic and repetitive ones.
To measure this, we compute the $n$-gram overlap between generated questions and all questions in the conversation history for all systems.
The lower this repetition is, the more likely the system is bringing up new entities or topics in its questions, and thus more specific to the given conversation history.
As can be seen in Figure \ref{fig:ngram_overlap},
our system improves upon the baseline system by reducing this repetition noticeably in longer $n$-grams ($n \ge 3$).
When $n$ is very large ($n\ge 8$), our pragmatic system is less repetitive even compared to the human reference, which often contains long and repetitive questions like \emph{Are there any other interesting aspects about this article?}\ as a generic inquiry for more information.

\diff{}{
\section{Human Evaluation Details} \label{sec:human_eval}

In this section, we include further details about how the human evaluation is carried out to compare different systems.

We begin by randomly sampling 200 turns of questions from the 7354 question-answer pairs in the test set, and collect the questions from the human reference, the baseline system, and our finetuned system.
Then, for each turn, we shuffle the order of the three candidate questions, and present them in a group to the annotators.
The questions are accompanied with the entity and topic under discussion, as well as the conversation history from the \quac{} dataset that led up to the turn under evaluation.
We ask the annotators to provide a ranking amongst the questions from these unidentified systems, and allow ties when the annotators cannot observe a qualitative difference between two or more question candidates.
An example annotation task for a turn of questions can be found in Figure~\ref{fig:annotation_interface}.
To encourage high inter-annotator agreement, we first conducted a trial annotation on 20 examples on the dev set, and composed annotation guidelines (see Figure \ref{fig:guidelines}) with some minimal examples to clarify edge cases.

\begin{figure}
    \centering
    \fbox{
    \includegraphics[width=0.45\textwidth]{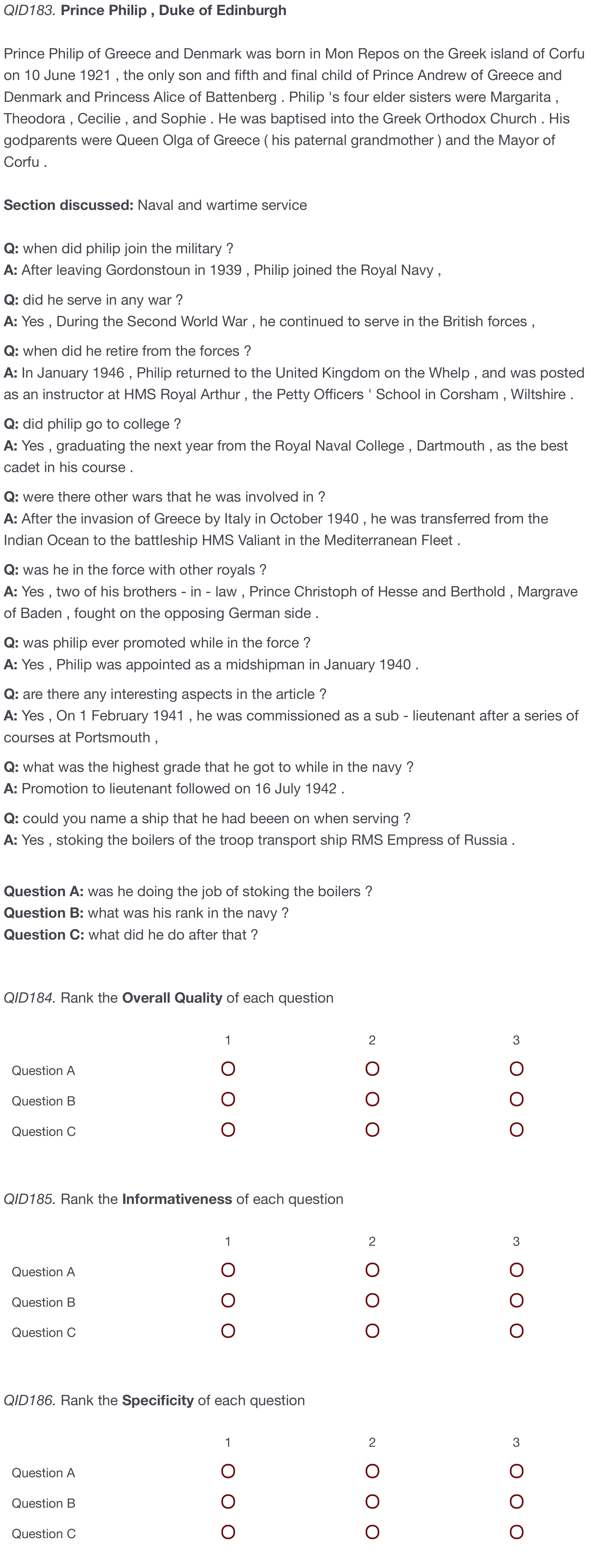}
    }
    \caption{The annotation interface for the human evaluation, which is built with Qualtrics.com.
    The annotators are given the title of the Wikipedia article under discussion, a short introductory paragraph, the title of the Wikipedia section discussed, the \quac{} conversation history that leads up to the current turn, and the candidate questions to evaluate.
    They are asked to rank these candidate questions on overall quality, informativeness, and specificity, as we outline in the guidelines.
    } \label{fig:annotation_interface}
\end{figure}

\begin{figure*}
    \centering
    \fbox{
    \includegraphics[width=0.75\textwidth]{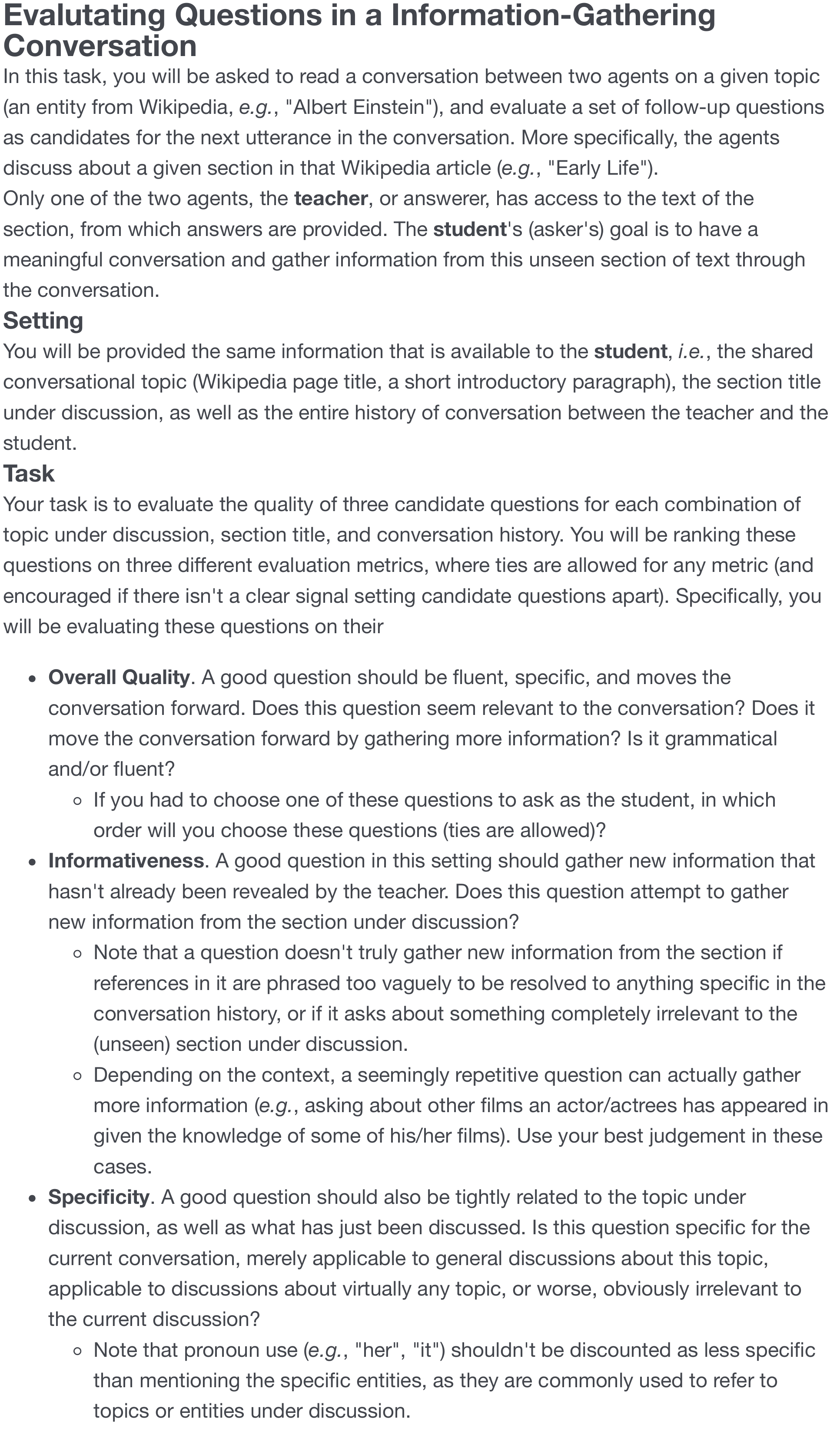}
    }
    \caption{Human evaluation guidelines to compare system-generated questions with the human reference.} \label{fig:guidelines}
\end{figure*}

}

\end{document}